\RequirePackage{fix-cm}
\documentclass[twocolumn]{svjour3}          
\smartqed  
\usepackage{url}
\usepackage{graphicx}
\usepackage{booktabs}
\usepackage{xcolor}
\usepackage{amsmath}

\journalname{Signal Image Video Processing}
\begin{document}

\title{On the Use of Deep Learning for Blind Image Quality Assessment}

\author{Simone Bianco \and Luigi Celona \and Paolo Napoletano \and Raimondo Schettini}


\institute{Dept. of Informatics, Systems and Communication, University of Milano-Bicocca, Viale Sarca 336, Milan 20126, Italy \\
              \email{ \{bianco,celona,napoletano,schettini\}@disco.unimib.it}
}

\date{Received: date / Accepted: date}

\maketitle

\begin{abstract}

In this work we investigate the use of deep learning for distortion-generic blind image quality assessment. We report on different design choices, ranging from the use of features extracted from pre-trained Convolutional Neural Networks (CNNs) as a generic image description, to the use of features extracted from a CNN fine-tuned for the image quality task. Our best proposal, named DeepBIQ, estimates the image quality by average-pooling the scores predicted on multiple sub-regions of the original image. 
Experimental results on the LIVE In the Wild Image Quality Challenge Database show that DeepBIQ outperforms the state-of-the-art methods compared, having a Linear Correlation Coefficient (LCC) with human subjective scores of almost 0.91. These results are further confirmed also on four benchmark databases of synthetically distorted images: LIVE, CSIQ, TID2008 and TID2013.


\end{abstract}
\section{Introduction}
\label{intro}
Digital pictures may have a low perceived visual quality. Capture settings, such as lighting, exposure, aperture, sensitivity to noise, and lens limitations, if not properly handled could cause annoying image artifacts that lead to an unsatisfactory perceived visual quality. Being able to automatically predict the quality of digital pictures can help to handle low quality images or to correct their quality during the capture process~\cite{bovik2013automatic}. An automatic image quality assessment (IQA) algorithm, given an input image, tries to predict its perceptual quality. The perceptual quality of an image is usually defined as the mean of the individual ratings of perceived quality assigned by human subjects ({Mean Opinion Score} - MOS).

In recent years, many IQA approaches have been proposed \cite{manap2015non,soundararajan2013survey}.
They can be divided into three groups, depending on the additional information needed: full-reference image quality assessment (FR-IQA) algorithms e.g. 
\cite{eckert1998perceptual,pappas2000perceptual,wang2004image,bianco2009image,he2011image,alaei2016image}, 
reduced-reference image quality assessment (RR-IQA) algorithms, and no-reference/blind image quality assessment (NR-IQA) algorithms e.g. \cite{moorthy2011blind,mittal2012no,mittal2013making,mahmoudpour2016no,li2016no,li2016no1}. FR-IQA algorithms perform a direct comparison between the image under test and a reference or original in a properly defined image space \cite{ciocca2014assess}. 
RR-IQA algorithms are designed to predict image quality with only partial information about the reference image \cite{ciocca2014assess}. In their general form, these methods extract a number of features from both the reference and the image under test, and image quality is assessed only by the similarity of these features. NR-IQA algorithms assume that image quality can be determined without a direct comparison between the original and the image under test \cite{ciocca2014assess}. Thus, they  can be used whenever the original image is unavailable. NR-IQA algorithms can be further classified into two main sub-groups: to the first group belong those targeted to estimate the presence of a specific image artifact (i.e. blur, blocking, grain, etc.) \cite{ciancio2011no,corchs2014no}; to the second group the ones that estimate the overall image quality and thus are distortion generic~\cite{mittal2015no,seshadrinathan2011automatic,bovik2013automatic,ciocca2014assess}. In this work we focus on distortion-generic NR-IQA. 

Most of the distortion-generic methods estimate the image quality by measuring deviations from Natural Scene Statistic (NSS) models~\cite{bovik2013automatic} that capture the statistical ``naturalness'' of non-distorted images. 
The Natural Image Quality Evaluator (NIQE) \cite{mittal2013making} is based on the construction of a quality aware collection of statistical features based on a space domain NSS model. 
The Distortion Identification-based Image Verity and INtegrity Evaluation (DIIVINE) index \cite{moorthy2011blind} is based on a two-stage framework for
estimating quality based on NSS models, involving distortion identification and distortion-specific quality assessment. 
C-DIIVINE \cite{zhang2014c} is an extension of the DIIVINE algorithm in the complex domain. 
The BLIINDS-II \cite{saad2012blind} method, given an input image, computes a set of features and then uses a Bayesian approach to predict quality scores. 

The Blind/Referenceless Image Spatial QUality Evaluator (BRISQUE) \cite{mittal2012no} operates in the spatial domain and is also based on a NSS model. 

The use of a database of images along with their subjective scores is fundamental for both the design and the evaluation of IQA algorithms \cite{sheikh2005live,Ghadiyaram2016}.
Recent approaches to the blind image quality assessment problem use these images coupled with the corresponding human provided quality scores within machine learning frameworks to learn directly from the data a quality measure. The Feature maps based Referenceless Image QUality Evaluation Engine (FRIQUEE) \cite{Ghadiyaram2014,Ghadiyaram2016} combines a deep belief net and a SVM to predict image quality. Tang et al. \cite{tang2014blind} define a simple radial basis function on the output of a deep belief network to predict the perceived image quality. 
Hou et al. \cite{hou2015blind} propose to represent images by NSS features and to train a discriminative deep model to classify the features into five grades (i.e. excellent, good, fair, poor, and bad). Quality pooling is then applied to convert the qualitative labels into scores. 
In \cite{lv2015difference} a model is proposed which uses local normalized multi-scale difference of Gaussian (DoG) response as feature vectors. Then, a three-steps framework based on a deep neural network is designed and employed as pooling strategy.
Ye et al. \cite{ye2013real} presented a supervised filter learning based algorithm that uses a small set of supervised learned filters and operates directly on raw image patches. 
Later they extended their work using a shallow convolutional neural network \cite{kang2014convolutional}.
The same CNN architecture has been then used to simultaneously estimate image quality and identify the distortion type \cite{kang2015simultaneous} on a single-type distortion dataset \cite{sheikh2005live}. Kottayil et al. \cite{kottayil2016color} used a hybrid approach composed by a shallow CNN architecture followed by a regressor to refine the quality score prediction


Features extracted from CNN pre-trained for object and scene recognition tasks, have been shown to provide image representations that are rich and highly effective for various computer vision tasks. This paper investigates their use for multiple generic distortions NR-IQA and their capability to model the complex dependency between image content and subjective image quality~\cite{allen2007image,triantaphillidou2007image,corchs2014no}.
The hypothesis motivating our research is that the presence of image distortion such as JPEG compression, noise, blur, etc. is captured and modelled by these features as well. Furthermore, the more concepts the CNN has been trained to recognize, the better are the extracted features. 
We evaluate the effect of several design choices: 
\begin{itemize}
\item[i)] the use of different features extracted from CNNs that are pre-trained on different image classification tasks for an increasing variety and number of concepts to recognize; 
\item[ii)] the use of a number of different image sub-regions (opposed to the use of the whole image) to better capture image artifacts that may be local or partially masked by specific image content;
\item[iii)] the use of different strategies for feature and score predictions pooling.  
\end{itemize}
We then propose a novel procedure for the fine-tuning of a CNN for multiple generic distortions NR-IQA, which consists in discriminatively fine-tuning the CNN to classify image crops into five distortion classes (i.e. bad, poor, fair, good, and excellent) and then using it as feature extractor.  Whatever is the feature extraction strategy and the related CNN, we finally exploit a Support Vector Regression (SVR) machine 
to learn the mapping function from the CNN features to the perceived quality scores \cite{li2016no}.


The experiments are conducted on the \emph{LIVE In the Wild Image Quality Challenge Database} which contains widely diverse authentic image distortions on a large number of images captured using a representative variety of modern mobile devices~\cite{ghadiyaram2014crowdsourced}. The result of this study is a CNN suitably adapted to the blind quality assessment task that accurately predicts the quality of images with a high agreement with respect to human subjective scores. Furthermore, we show the applicability of our method to the legacy LIVE Image Quality Assessment Database \cite{sheikh2005live}, CSIQ \cite{larson2010most}, TID2008 \cite{ponomarenko2009tid2008} and TID2013 \cite{ponomarenko2013color}.


\section{Deep Learning for BIQ assessment}

Deep Convolutional Neural Networks (CNNs) are a class of learnable architectures used in many image domains \cite{lecun2015deep,razavian2014cnn} such as recognition, annotation, retrieval, object detection, etc. CNNs are usually composed of several layers of processing, each involving linear as well as non-linear operators that are jointly learned in an end-to-end manner to solve a particular task. 

A typical CNN architecture consists of a set of stacked layers: convolutional layers to extract local features; point-wise non-linear mappings; pooling layers, which aggregates the statistics of the features at nearby locations; and fully connected layers. The result of the last fully connected layer is the CNN output. CNN architectures vary in the number of layers, the number of outputs per layer, the size of the convolutional filters, and the size and type of spatial pooling. CNNs are usually trained in a supervised manner by means of standard back-propagation \cite{lecun2012efficient}.

In practice, very few people train an entire CNN from scratch, because it is relatively rare to have a dataset of sufficient size. Instead, it is common to take a CNN that is pre-trained on a different large dataset (e.g. ImageNet \cite{deng2012imagenet}), and then use it either as a feature extractor or as an initialization for a further learning process (i.e. transfer learning, known also as fine-tuning \cite{bengio2012deep}). 
Among possible CNN architectures~\cite{jia2014caffe,simonyan2014very,szegedy2015going}, after preliminary investigations, we have chosen the Caffe network architecture \cite{jia2014caffe} (inspired by the AlexNet \cite{krizhevsky2012imagenet}) as a feature extractor on top of which we exploit a Support Vector Regression (SVR) machine 
with a linear kernel to learn a mapping function from the CNN features to the perceived quality scores (i.e. MOS). 
The detailed architecture of the CNN used is reported in Table \ref{CNNarch}.

\begin{table*}[]
\scriptsize
\centering
\caption{Architecture of Caffe network. 
}
\label{CNNarch}
\resizebox{0.85\textwidth}{!}{
\begin{tabular}{lccccccccccccc}
\toprule
			& \it{conv1}	& \it{pool1}		& \it{norm1}		& \it{conv2}		& \it{pool2}      
			& \it{norm2} 	& \it{conv3}		& \it{conv4}		& \it{conv5}		& \it{pool5}
			& \it{fc6}		& \it{fc7}			& \it{fc8}  		\\ 
\midrule
Type		& Conv	& MaxPool	& LRN		& Conv		& MaxPool
			& LRN	& Conv		& Conv		& Conv		& MaxPool
			& FC	& FC		& FC			\\ 
Kernel size	& $11\times11$	&	$3\times3$	&			&	$5\times5$	
& $3\times3$	&       			&	$3\times3$	& $3\times3$	& 	$3\times3$
& $3\times3$	&      			&      			& \\ 
Depth		& 96	&	&	& 256	&	&	& 384	& 384	& 256	&		& 4096	& 4096	& \\ 
Stride		& 4		& 2	&	& 1		& 2	&	& 1		& 1		& 1		& 2		&		&		& \\
Padding		& 0		&	&	& 2		&	&	& 1		& 1		& 1		&		&		&		& \\
\bottomrule
\end{tabular}}
\end{table*}

Given an input image, the CNN performs all the multilayered operations and the corresponding feature vector is obtained by removing the final softmax nonlinearity and the last fully-connected layer. The length of the feature vector is 4096. 

In this work we evaluate the effect of several design choices for feature extraction, such as: i) the use of different CNNs that are pre-trained on different image classification tasks; ii) the use of a number of different image sub-regions (opposed to the use of the whole image) as well as the use of different strategies for feature and score prediction pooling; iii) the use of a CNN that is fine-tuned for category-based image quality assessment.


\subsection{Image description using pre-trained CNNs}
\label{sec:pretrainedCNN}
Razavian et al. \cite{razavian2014cnn} showed that the generic descriptors extracted from convolutional neural networks are very powerful and their use outperforms hand crafted, state-of-the-art systems in many visual classification tasks. Within the approach 
previously described, 
our baseline consists in the use of off-the-shelf CNNs as feature extractors. Features are computed by feeding the CNN with the whole image, that must be resized to fit its predefined input size.
We experiment with the use of three CNNs sharing the same architecture that have been pre-trained on three different image classification tasks:
\begin{enumerate}
\item[-] ImageNet-CNN, which has been trained on 1.2 million images of ImageNet (ILSVRC 2012) for object recognition belonging to 1,000 categories;
\item[-] Places-CNN, which has been trained on ~2.5 million images of the Places Database for scene recognition belonging to 205 categories;
\item[-]  ImageNet+Places-CNN, which has been trained using 3.5 million images from 1,183 categories, obtained by merging the scene categories from Places Database and the object categories from ImageNet.
\end{enumerate}

\subsection{Feature and prediction pooling strategies}
In the design choice described in Section \ref{sec:pretrainedCNN}, we resized the image to match the predefined CNN input size. Since the resizing operation can mask some image artifacts, we consider here a different design choice in which CNN features are computed on multiple sub-regions (i.e. crops) of the input image. Crops dimensions are chosen to be equal to the CNN input size so that no scaling operation is involved. Each crop covers almost 21\% of the original image (227$\times$227 out of 500$\times$500 pixels), thus the use of multiple crops permits to evaluate the local quality. The final image quality is then computed by pooling the evaluation of each single crop. This permits, for instance, to distinguish between a globally blurred image and a high-quality depth-of-field image.

We experiment the use of a different number randomly selected sub-regions \cite{krizhevsky2012imagenet}, ranging from 5 to 50.  The information coming from the multiple crops has to be fused to predict a single quality score for the whole image. The different fusion strategies are here reported:
\begin{enumerate}
\item[-] \emph{feature pooling}: information fusion is performed element by element on the sub-region feature vectors to generate a single feature vector for each image 
 minimum, average, and maximum feature pooling are considered. 
\item[-] \emph{feature concatenation}: information fusion is performed by concatenating the sub-region feature vectors in a single longer feature vector. 
\item[-] \emph{prediction pooling}: information fusion is performed on the predicted quality scores. The SVR predicts a quality score for each image crop, and these scores are then fused using a minimum, average, or maximum pooling operators. 
\end{enumerate}



\subsection{Image description using a fine-tuned CNN}
Convolutional neural networks usually require millions of training samples in order to avoid overfitting. Since in the blind image quality assessment domain the amount of data available is not so large, we investigate the fine-tuning of a pre-trained CNN exploiting the available NR-IQA data. When the amount of data is small, it is likely best to keep some of the earlier layers fixed and only fine-tune some higher-level portion of the network. This procedure, which is also called transfer learning~\cite{bengio2012deep}, is feasible since the first layers of CNNs learn features similar to Gabor filters and color blobs that appear not to be specific to a particular image domain; while the following layers of CNNs become progressively more specific to the given domain~\cite{bengio2012deep}.

We start the fine-tuning procedure to the image quality assesment task by substituting the last fully connected layer of a pre-trained CNN with a new one initialized with random values. The new layer is trained from scratch, and the weights of the other layers are updated using the back-propagation algorithm \cite{lecun2012efficient} with the available data for image quality assessment. In this work, image quality data are a set of images having  human average quality scores (i.e. MOS).
The CNN is discriminatively fine-tuned to classify image sub-regions into five classes 
according to the 5-points MOS scale. 
The five classes are obtained by a crisp partition of the MOS: bad (score $\in \ [0, 20]$), poor (score $\in \ ]20,40]$), fair (score $\in \ ]40,60]$), good (score $\in \ ]60,80]$), and excellent (score $\in \ ]80,100]$).
Once the CNN is trained, it is used for feature extraction
, just like one of the pre-trained CNNs.

\section{Experimental results}
Different standard databases are available to test the algorithms’ performance with respect to the human subjective judgments. Most of them have been created starting from high-quality images, and adding synthetic distortions. However, as pointed out by  Ghadiyaram and Bovik \cite{ghadiyaram2014crowdsourced}: ``images captured using typical real-world mobile camera devices are usually afflicted by complex mixtures
of multiple distortions, which are not necessarily well-modelled by the synthetic distortions found in existing databases''.

\begin{figure}[tb]
\centering
\resizebox{1.0\columnwidth}{!}{
\begin{tabular}{ccccc}
\includegraphics[width=2.5truecm]{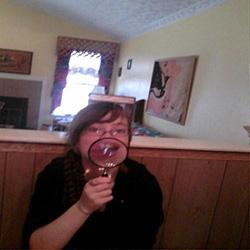} &
\includegraphics[width=2.5truecm]{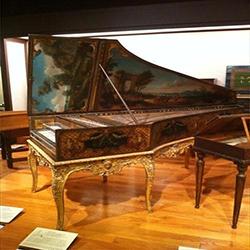} &
\includegraphics[width=2.5truecm]{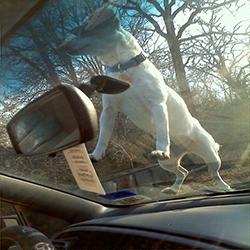} &
\includegraphics[width=2.5truecm]{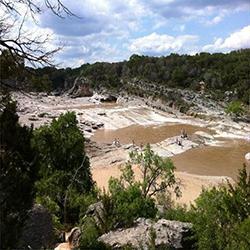} &
\includegraphics[width=2.5truecm]{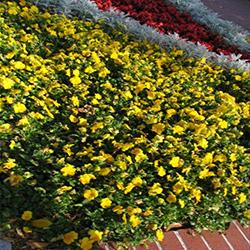} \\
MOS= 25.3 & MOS= 77.2 & MOS= 64.7 & MOS= 73.0 & MOS= 54.4 \\
\includegraphics[width=2.5truecm]{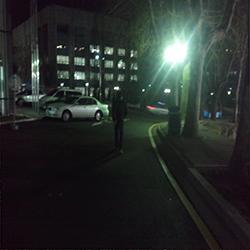} &
\includegraphics[width=2.5truecm]{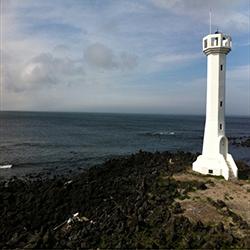} &
\includegraphics[width=2.5truecm]{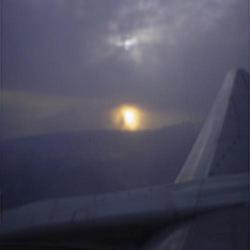} &
\includegraphics[width=2.5truecm]{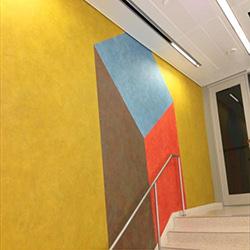} &
\includegraphics[width=2.5truecm]{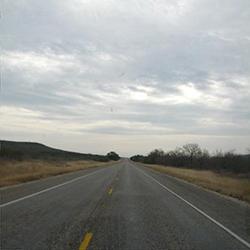} \\
 MOS= 24.5 & MOS= 81.8 & MOS= 16.1 & MOS= 78.4 & MOS= 64.0 \\
\end{tabular}
}
\caption{Examples from the LIVE In the Wild IQ Chall.DB.}
\label{LBIQCdb_samples}
\end{figure}

We evaluate the different design choices within the proposed approach on the LIVE In the Wild Image Quality Challenge Database
\cite{ghadiyaram2014crowdsourced,Ghadiyaram2016}. It contains 1,162 images with resolution equal to $500\times500$ pixels affected by diverse authentic distortions and genuine artifacts such as low-light noise and blur, motion-induced blur, over and underexposure, compression errors, etc. Figure \ref{LBIQCdb_samples} shows some database samples. Database images have been rated by many thousands of subjects via an online crowdsourcing system designed for subjective quality assessment. Over 350,000 opinion scores from over 8,100 unique human observers have been gathered. The mean opinion score (MOS) of each image is computed by averaging the individual ratings across subjects, and used as ground truth quality score. The MOS values are in the [1, 100] range.

We compared the different design choices within the proposed approach with a number of leading blind IQA algorithms. Since most of these algorithms are machine learning-based training procedures, following \cite{Ghadiyaram2016} in all the experiments we randomly split the data into 80\% training and 20\% testing sets, using the training data to learn the model and validating its performance on the test data. To mitigate any bias due to the division of data, the random split of the dataset is repeated 10 times. For each repetition we compute the Pearson’s Linear Correlation Coefficient (LCC) and the Spearman’s Rank Ordered Correlation Coefficient (SROCC) between the predicted and the ground truth quality scores, reporting the median of these correlation coefficients across the 10 splits. In all the experiments we use the Caffe open-source framework \cite{jia2014caffe} for CNN training and feature extraction, and the LIBLINEAR library \cite{fan2008liblinear} for SVR training.

\subsection{Experiment I: pre-trained CNNs}
We extract the 4096-dimensional features from the \textit{fc7} layer of the pre-trained ImageNet-CNN, Places-CNN and ImageNet+Places-CNN. Since these CNNs require an input with a dimensionality equal to $227\times227$ pixels, we rescale the original $500\times500$ images to $256\times256$ keeping aspect ratio, and then we crop out the central $227\times227$ sub-region from the resulting image. All the images are pre-processed by subtracting the mean image, that is computed by averaging all the images in the training set on which the CNN was pre-trained. The median LCC and SROCC over the 10 train-test splits are reported in Table \ref{Exp1_results}. From the results it is possible to see that ImageNet+Places-CNN outperforms both Imagenet-CNN and Places-CNN, with Places-CNN giving the worst performance confirming our original hypothesis that the more concept the CNN has been trained to recognize, the more effective are its features for modeling generic image content.

\begin{table}
\scriptsize

\centering
\caption{Median LCC and SROCC across 10 train-test random splits of the LIVE In the Wild Image Quality Challenge Database considering only the central crop of the subsampled image as input for the pre-trained CNNs considered.}
\label{Exp1_results}
\begin{tabular}{lcc}
\toprule
    			& \multicolumn{1}{c}{LCC}	& \multicolumn{1}{c}{SROCC} 	\\	
\midrule
Imagenet-CNN	& 	0.6782					&	0.6381 						\\
Places-CNN		&	0.6267					& 	0.6055						\\
ImageNet+Places-CNN		& 	\textbf{0.7215} 		&	\textbf{0.7021} 			\\ 
\bottomrule
\end{tabular}
\end{table}

\subsection{Experiment II: feature and prediction pooling}
In the previous experiment the resize operation could have reduced the effect of some artifacts, e.g. noise. In order to keep unchanged the distortion level we evaluate the performances of features extracted from a variable number of randomly cropped $227\times227$ sub-regions from the original image. This choice is confirmed in preliminary experiments (not reported here due to lack of space) where  taking crops at different scales demonstrated to perform worse than taking them at the original image scale. 

Given the results in the previous section, the only features considered here are those extracted using the ImageNet+Places-CNN.

We evaluate three different fusion schemes for combining the information generated by the multiple sub-regions to obtain a single score prediction for the whole image. 
The first scheme is feature pooling that can be seen as an early fusion approach, performing element-wise fusion on the feature vectors. The second scheme is feature concatenation, performing information fusion by concatenating the multiple feature vectors into a single feature vector. The third scheme is prediction pooling that can be seen as a late fusion approach, where information fusion is performed on the predicted quality scores.

In all the experiments the number of random crops is varied between 5 and 50 in steps of 5. 
The numerical values of LCC and SROCC for the best configurations of each fusion scheme (across pooling operators and number of crops) are reported in Table \ref{Exp2_table_results}. The optimal number of crops has been selected by running the two-sample $t-$test whose results are reported as additional material.
Concerning the best configurations reported in Table \ref{Exp2_table_results}, the output of the two-sample $t-$test  shows that the results obtained by feature average-pooling are statistically better than both those obtained by feature concatenation ($p$-value equal to 3.4$\cdot 10^{-9}$) and prediction average-pooling ($p-$value equal to 8.8$\cdot 10^{-5}$). The difference between feature concatenation and prediction average-pooling is not significative instead ($p-$value equal to 0.23).

\begin{table}[]
\centering
\scriptsize

\caption{Median LCC and SROCC across 10 train-test random splits of the LIVE In the Wild IQ Chall. DB considering randomly selected crops as input for the ImageNet+Places-CNN and three different fusion approaches.}
\label{Exp2_table_results}
\begin{tabular}{lcc}
\toprule
                                      	& \multicolumn{1}{c}{LCC} & \multicolumn{1}{c}{SROCC} \\ 
\midrule                                                                          
Feature pooling (avg-pool,@30crops)   	& \textbf{0.7938}           & \textbf{0.7828}             \\
Feature concatenation (@35crops)        & 0.7864          & 0.7724         \\ 
Prediction pooling (avg-pool,@20crops)	& 0.7873           & 0.7685            \\
\bottomrule
\end{tabular}
\end{table}


\subsection{Experiment III: fine-tuned CNN}
In all previous experiments we use pre-trained CNNs for feature extraction. In this experiment instead, we fine-tune the ImageNet+Places-CNN for the NR-IQA task. 
The CNN is discriminatively fine-tuned to classify image crops into five distortion classes (i.e. bad, poor, fair, good, and excellent) obtained by crisp partitioning the MOS into five disjoint sets. 
Since the number of images belonging to the five sets is uneven
, during training we use a sample weighting approach \cite{huang2005weighted} giving larger weights to images belonging to less represented distortion classes \cite{zhou2006training}. Weights are computed as the ratio between the frequency of the most represented class and the frequency of the class to which the image belongs. 

On the NR-IQA task this weighting scheme gives better results compared to batch-balancing (i.e. assuring that in each batch all the classes are evenly sampled) since it guarantees more heterogeneous batches.

Given the results of the previous experiments, we only evaluate the performance of the fine-tuned CNN with feature pooling and prediction pooling with the average operator. We fine-tune the network for 5,000 iterations using Caffe framework \cite{jia2014caffe} on a NVIDIA K80 GPU. The total training time was about 2 hours, while predicting the MOS for a single image at test time requires about 20ms.

The numerical values of LCC and SROCC for the best configurations are reported in Table \ref{Exp4_table_results}. 
{As for the previous experiment, the optimal number of crops has been selected by running the two-sample $t-$test test whose results are reported as additional material.} 
Concerning the best configurations reported in Table \ref{Exp4_table_results}, the output of the two-sample $t-$test shows that the results obtained by prediction average-pooling are statistically better than those obtained by feature average-pooling ($p$-value equal to 4.7$\cdot 10^{-4}$).
%

\begin{table}[]
\centering
\scriptsize

\caption{Median LCC and SROCC across 10 train-test random splits of the  LIVE In the Wild
Image Quality Challenge Database considering randomly selected crops as input for the fine-tuned CNN and two different fusion approaches.}
\label{Exp4_table_results}
\begin{tabular}{lcc}
\toprule
                                        & \multicolumn{1}{c}{LCC}	& \multicolumn{1}{c}{SROCC}	\\ 
\midrule
Feature pooling (avg-pool,@20crops)   	& 0.9026      				&	0.8851					\\ 
Prediction pooling (avg-pool,@25crops)	& \textbf{0.9082}			&	\textbf{0.8894}			\\
\bottomrule
\end{tabular}
\end{table}

In Table \ref{results} we compare the results of the different instances of the proposed approach, that we name DeepBIQ, with those of some NR-IQA algorithms in the state of the art. From the results it is possible to see that the use of a pre-trained CNN on the whole image is able to give slightly better results than the best in the state of the art. The use of multiple crops with average-pooled features is able to improve LCC and SROCC with respect to the best method in the state of the art by 0.08 and 0.11 respectively. Finally the use of the fine-tuned CNN with multiple image crops and average-pooled predictions is able to improve LCC and SROCC by 0.20 and 0.21 respectively. 
Since the MOS is assumed to be the ground truth metric, we also report performance in terms of MOS statistics: the ground truth MOS is predicted with an RMSE of 8.59\% and a MAE of 6.42\%.


Error statistics  may not give an intuitive idea of how well a NR-IQA algorithm performs. On the other hand, individual human scores can be rather noisy. 
Taking into account that the LIVE In the Wild Image Quality Challenge Database gives for each image the MOS as well as the standard deviation of the human subjective scores, to have an intuitive assessment of DeepBIQ performance we proceed as follows: we divide the absolute prediction error of each image by the standard deviation of the subjective scores for that particular image. We then build a cumulative histogram and collect statistics at one, two, and three standard deviations. Results indicate that 97.2\% of our predictions are below $\sigma$, 99.4\% below $2\sigma$ and 99.8\% below $3\sigma$. Assuming a normal error distribution, this means that in most of the cases the image quality predictions made by DeepBIQ are closer to the average observer than those of a generic human observer.


\begin{table}
\centering
\scriptsize
\caption{Median LCC and median SROCC across 10 train-test random splits of the LIVE In the Wild IQ Chall. DB.}
\label{results}
\begin{tabular}{lcc}
\toprule
									& \multicolumn{1}{c}{LCC} & \multicolumn{1}{c}{SROCC} \\ 
\midrule
DIIVINE \cite{moorthy2011blind} 													& 0.56 		& 0.51		\\
BRISQUE \cite{mittal2012no}															& 0.61 		& 0.60	 	\\
BLIINDS-II \cite{saad2012blind}														& 0.45 		& 0.40	 	\\
S3 index \cite{vu2012spectral}														& 0.32 		& 0.31 		\\
NIQE \cite{mittal2013making}														& 0.48 		& 0.42		\\
C-DIIVINE \cite{zhang2014c}															& 0.66 		& 0.63		\\
FRIQUEE \cite{Ghadiyaram2014,Ghadiyaram2016}										& 0.71 		& 0.68	 	\\
HOSA \cite{xu2016blind}                   											& 	- 		& 0.65 		\\
\textbf{DeepBIQ (Exp. I)}								& 0.72		& 0.70		\\	
\textbf{DeepBIQ (Exp. II)}		& 0.79		& 0.79 		\\
\textbf{DeepBIQ (Exp. III)}		& \bf{0.91}	& \bf{0.89}	\\
\bottomrule
\end{tabular}
\end{table}

For sake of comparison with other methods in the state of the art, as an additional experiment we evaluate our method on the older but widely used benchmark databases of synthetically distorted images:
LIVE Image Quality Assessment Database\cite{sheikh2005live}, Categorical Subjective Image Quality (CSIQ) Database \cite{larson2010most},  TID2008 \cite{ponomarenko2009tid2008},  TID2013 \cite{ponomarenko2013color}.

We evaluate our method on these datasets dealing with the different human judgements and distortion ranges by only re-training the SVR, while keeping the CNN unchanged. We follow the experimetal protocol used in \cite{kang2014convolutional,kang2015simultaneous}. This protocol consists in running 100 iterations, where in each iteration 60\% of the reference images and their distorted versions is randomly select as the training set, 20\% as the validation set, and the remaining 20\% as the test set. The experimental results in terms of average LCC and SROCC values on LIVE are reported in Table \ref{tab:live2}, on CSIQ in Table \ref{tab:csiq}, on TID2008 in Table \ref{tab:tid2008}, and on TID2013 in Table \ref{tab:tid2013}. 
From these results it is possible to see that our method, DeepBIQ, is able to obtain the best performance in terms of both LCC and SROCC notwithstanding that differently from all the other methods reported, the features have been learned on a different dataset containing images with real distortions and not on a portion of the test database itself. Therefore, the results confirm the effectiveness of our approach for no-reference IQ assessment.

\begin{table}
\centering
\scriptsize
\caption{Median LCC and median SROCC across 100 random splits of the legacy LIVE Image Quality Assessment DB.}
\label{tab:live2}
\begin{tabular}{lcc}
\toprule
Method & LCC & SROCC \\
\midrule
DIIVINE \cite{moorthy2011blind} 				& 0.93 &   0.92 	\\
BRISQUE \cite{mittal2012no}						& 0.94 &   0.94 	\\
BLIINDS-II \cite{saad2012blind}					& 0.92 &   0.91 	\\
NIQE \cite{mittal2013making}					& 0.92 &   0.91 	\\
C-DIIVINE \cite{zhang2014c}						& 0.95 &   0.94 	\\
FRIQUEE \cite{Ghadiyaram2014,Ghadiyaram2016}  	& 0.95 & 0.93 \\
ShearletIQM \cite{mahmoudpour2016no} & 0.94 & 0.93 \\
MGMSD \cite{alaei2016image} & 0.97 & \bf{0.97}\\
Low Level Features \cite{kottayil2016color} & 0.95 & 0.94\\
Rectifier Neural Network \cite{tang2014blind} 	& --   & 0.96 \\
Multi-task CNN \cite{kang2015simultaneous}    	& 0.95 & 0.95 \\
Shallow CNN	\cite{kang2014convolutional}		& 0.95 & 0.96 \\
DLIQA \cite{hou2015blind}						& 0.93 & 0.93 \\
HOSA \cite{xu2016blind} 						& 0.95 & 0.95 \\
CNN-Prewitt \cite{li2016no1} & 0.97 & 0.96\\
CNN-SVR \cite{li2016no} & 0.97 & 0.96\\
\bf{DeepBIQ} 									& \bf{0.98} & \bf{0.97}\\
\bottomrule
\end{tabular}
\end{table}

\begin{table}
\centering
\scriptsize
\caption{Median LCC and median SROCC across 100 train-val-test random splits of the CSIQ.}
\label{tab:csiq}
\begin{tabular}{lcc}
\toprule
Method & LCC & SROCC \\
\midrule
DIIVINE \cite{moorthy2011blind} 				& 0.90 & 0.88 	\\
BRISQUE \cite{mittal2012no}						& 0.93 & 0.91 	\\
BLIINDS-II \cite{saad2012blind}					& 0.93 & 0.91 	\\
Low Level Features \cite{kottayil2016color} & 0.94 & 0.94\\
Multi-task CNN \cite{kang2015simultaneous}    	& 0.93 & 0.94 		\\
HOSA \cite{xu2016blind} 						& 0.95 & 0.93 		\\
\bf{DeepBIQ} 									& \bf{0.97} & \bf{0.96}\\
\bottomrule
\end{tabular}
\end{table}

\begin{table}
\centering
\scriptsize
\caption{Median LCC and median SROCC across 100 train-val-test random splits of the TID2008.}
\label{tab:tid2008}
\begin{tabular}{lcc}
\toprule
Method & LCC & SROCC \\
\midrule
DIIVINE \cite{moorthy2011blind} 				& 0.90 &   0.88 	\\
BRISQUE \cite{mittal2012no}						& 0.93 &   0.91 	\\
BLIINDS-II \cite{saad2012blind}					& 0.92 &   0.90 	\\
MGMSD \cite{alaei2016image} & 0.88 & 0.89\\
Low Level Features \cite{kottayil2016color} & 0.89 & 0.88\\
Multi-task CNN \cite{kang2015simultaneous}    & 0.90  & 0.91 \\
Shallow CNN	\cite{kang2014convolutional}			  & 0.90  & 0.92 \\
\bf{DeepBIQ} & \bf{0.95} & \bf{0.95}\\
\bottomrule
\end{tabular}
\end{table}

\begin{table}
\centering
\scriptsize
\caption{Median LCC and median SROCC across 100 train-val-test random splits of the TID2013.}
\label{tab:tid2013}
\begin{tabular}{lcc}
\toprule
Method & LCC & SROCC \\
\midrule
DIIVINE \cite{moorthy2011blind} 				& 0.89 &   0.88 	\\	
BRISQUE \cite{mittal2012no}						& 0.92 &   0.89 	\\	
BLIINDS-II \cite{saad2012blind}					& 0.91 &   0.88 	\\	
Low Level Features \cite{kottayil2016color} & 0.89 & 0.88\\
HOSA \cite{xu2016blind} & \bf{0.96}  & 0.95 \\
\bf{DeepBIQ} & \bf{0.96} & \bf{0.96}\\
\bottomrule
\end{tabular}
\end{table}

\section{Conclusions}
In this work we have investigated the use of deep learning for distortion-generic blind image quality assessment. We report on different design choices in three different experiments, ranging from the use of features extracted from pre-trained Convolutional Neural Networks (CNNs) as a generic image description, to the use of features extracted from a CNN fine-tuned for the image quality task.

Our best proposal, named DeepBIQ, consists of a CNN originally trained to discriminate 1,183 visual categories that is fine-tuned for category-based image quality assessment. This CNN is then used to exctract features that are then fed to a SVR to predict the image quality score. By considering multiple image crops and exploiting the prediction pooling fusion scheme with the average operator, DeepBIQ reaches a LCC of almost 0.91, that is 0.20 higher than the best solution in the state of the art \cite{Ghadiyaram2016}. Furthermore, in many cases, the quality score predictions of our method are closer to the average observer than those of a generic human observer.
 
DeepBIQ is then further tested on four benchmark databases of synthetically distorted images: LIVE, CSIQ, TID2008 and TID2013. To deal with the different types of human opinion scores and distortion ranges, we only re-trained the SVR, while keeping the CNN unchanged. Experimental results show that DeepBIQ is able to outperform all the methods in the state of the art also on these datasets, even if the features have been learned on a different dataset, confirming the effectiveness of our approach for no-reference image quality assessment.

A web demo of the DeepBIQ network and additional materials are available at \url{http://www.ivl.disco.unimib.it/activities/deep-image-quality/}.

\bibliographystyle{spmpsci}      

\bibliography{refss}

%
%

\end{document}